\documentclass{ISMA_USD}

\usepackage{calrsfs}
\DeclareMathAlphabet{\pazocal}{OMS}{zplm}{m}{n}

\usepackage{tikz}
\usetikzlibrary{shapes,positioning,intersections,quotes,bayesnet}
\usetikzlibrary{decorations.pathmorphing}
\usetikzlibrary{decorations.markings}
\usetikzlibrary{arrows.meta,bending}

\hypersetup{pdftitle  = {Towards a GraphNARX model for structural dynamics},
	pdfauthor = {G. Tsialiamanis, K. Worden},
	pdfkeywords = {ISMA2026, USD2026, template}}

\title{Towards a GraphNARX model for structural dynamics}

\author[1]{G. Tsialiamanis}
\author[1]{K. Worden}
\affil[1]	{Dynamics Research Group, School of Mechanical, Aerospace, and Civil Engineering, University of Sheffield, Sheffield, Mappin St, Sheffield, United Kingdom \NewAffil}
		
\date{}

\begin{document}
%\bstctlcite{IEEEtran:BSTadapt}	

%%%%%%%%%%%%
% Abstract %
%%%%%%%%%%%%

% Enter your abstract in the command below.

\abstract{The current paper introduces the GraphNARX model, a graph-based nonlinear autoregressive exogenous model for structural dynamics. The model is built to account for the geometry of structures and to model dynamic behaviour as the interaction between elements of the structure. Similar to other types of NARX models, GraphNARX is used for system identification. Nodal attributes of the graph encode past displacements and nodal forces, and the output of the model is the next-timestep displacement. The graph formulation of the model allows it to generalise to structures of different dimensions and degrees of freedom. Moreover, because graph-learning is performed locally between the nodes, the GraphNARX model can also learn to infer displacements of nodes for which data are not available, i.e.\ to perform virtual sensing. The proposed model's performance is evaluated in a system-identification framework and a virtual-sensing framework, and achieves good accuracy and exhibits generalisation capabilities to structures of different sizes.}

\maketitle

%%%%%%%%%%%%%%%%%%%%%%%%%%
% Beginning of the paper %
%%%%%%%%%%%%%%%%%%%%%%%%%%

\section{Introduction}

Modelling physical systems, and particularly structural systems, has been the objective of academic research for many years. A lot of effort has been put into accurate prediction of behaviour of structures -- especially during the design phase -- in order to predict and avoid catastrophic situations. In different cases, particular behaviours of structures can be exploited and lead to innovative designs which solve existing problems. As a result, modelling approaches to under-design structural systems have progressed quite a lot, with, arguably, the most successful method being the \textit{finite element} (FE) method \cite{bathe2006finite}. Although design of structures makes such methods imperative for engineering practice, the inverse procedure can often be extremely useful as well.

The inverse problem, that of creating a model which accurately replicates an existing structure (or system more generally), is often referred to as \textit{system identification} \cite{noel2017nonlinear}. Such a process requires the acquisition of data from the physical structure. Identifying structures may range from calibrating a physical model, e.g.\ an FE model \cite{kiran2025state}, to fitting machine-learning models to data from structural systems \cite{liu2021machine} or hybrid approaches combining physics-based approaches with machine learning \cite{rogers2017grey}.

The two approaches, physics-based and data-driven, have their advantages and disadvantages. On the one hand, a physics-based approach can be more robust and, if the actual physics of the system are encoded into the model, generalisation outside the calibration domain is plausible. Examples of such systems and modelling approaches can be found in \cite{wigren2013three}. On the other hand, a data-driven approach, such as a neural network model offers the ability to learn the physics exclusively from data, however, machine learning models are mainly capable of interpolation, i.e.\ performing inference in situations similar to the situations used for training, which can hinder their generalisation performance. 

The current work is motivated by grey-box modelling \cite{rogers2017grey} and presents a graph-based methodology for structural identification in order to account for the geometry of structures (a major physical property) via the graph formulation, and attempt a form of extrapolation between structures of different sizes. The approach is based on the use of \textit{graph neural networks} (GNNs) \cite{battaglia2018relational} and \textit{nonlinear autoregressive exogenous models} (NARX) \cite{billings2013nonlinear}. The main aim of the current work is to present a system-identification method which is scalable in size. Moreover, because of the local learning that the graph-based method is based on, the methodology is evaluated in a virtual-sensing framework, i.e.\ inferring the displacements of points on the structure at which no sensor is placed. The structure of the paper is the following. In Section \ref{sec:graphnarx}, the formulation of the GraphNARX model is presented. In Section \ref{sec:sid_app}, an application of the GraphNARX model is presented for a simulated structure and its extrapolation capabilities to a larger structure are evaluated. In Section \ref{sec:VS_graphnarx}, the use of the GraphNARX model for virtual sensing is evaluated. Finally, in Section \ref{sec:discussion}, the results are discussed and conclusions are drawn.

\section{The GraphNARX model}
\label{sec:graphnarx}

\subsection{Nonlinear autoregressive exogenous (NARX) modelling}
\label{sec:narx}

Autoregressive exogenous (ARX) models are based on the assumption that the state of a dynamic system in the future is a function of past states of the system (autoregressive) and the external input to the system (exogenous). Such models refer to linear transition models from the past states to the future ones \cite{kopsaftopoulos2012vector}. Such an assumption may yield satisfactory results for a range of systems (or structures), however, it is quite restrictive, given how common nonlinearity is in nature \cite{worden2019nonlinearity, kerschen2006past}. Adding nonlinearity to ARX models, i.e.\ allowing for the transition function to be nonlinear, NARX models are created. Such models are defined in discrete time and can be generally described by,
\begin{equation}
    \bm{y}_{t+1} = f(\bm{y}_{t}, \bm{y}_{t-1}..., \bm{y}_{t-n_{y}}, \bm{x}_{t}, \bm{x}_{t-1}..., \bm{x}_{t-n_{x}})
    \label{eq:general_narx}
\end{equation}
where $\bm{y}_{t}$ is the vector describing the state of the structure at timestep $t$, $\bm{x}_{t}$ is the exogenous input to the system at timestep $t$, $f$ is the NARX model describing the transition from previous states to the future state $t+1$, $n_{y}$ is the number of previous state observations used by the model, and $n_{x}$ is the number of previous exogenous inputs to the system. The two numbers $n_{x}$ and $n_{y}$ are often referred to as the \textit{lags}.   

The simplest form that the function $f$ can take is that of a multinomial. Such a model is the \textit{polynomial-NARX} model, or \textit{p-NARX}. The model then takes the form of a polynomial of $n$ order of the state $y_{t}$. Note that if the state is a vector, which is the most common case, then the number of polynomial terms and cross terms grows enormously as the order of the polynomial is increased. Although the p-NARX model may be simple in its formalisation, it is a very useful tool for dynamic modelling and system identification. The coefficients of different orders of the past states reveal information about the system's nonlinear behaviour. For example, identifying a \textit{Duffing oscillator} using a p-NARX model is expected to yield cubic terms in the polynomial model. Furthermore, given a differential equation for a system, a correspondence between coefficients of the polynomial and physical parameters of the system can be found; thus translating the coefficients into meaningful quantities. 

A different approach to defining NARX models is to use more flexible and black-box models. Polynomials can approximate an underlying relationship arbitrarily accurately with enough terms; however, choosing their order can be difficult. Defining the coefficients of the polynomial may also be non-trivial. Especially if one thinks that a monomial of order $n$ can have quite similar behaviour with a monomial of order $2n, 3n$ etc, identifying the coefficients of each monomial can have a leakage effect leading to unstable or erroneous solutions. As a result, more flexible approximators have been used as the model $f$ in equation (\ref{eq:general_narx}).

A quite effective universal approximator, which has been used as a NARX model, is the \textit{Gaussian process} (GP) \cite{williams2006gaussian}. The NARX model which uses GPs as the transition relationship is the \textit{GP-NARX} model \cite{worden2014gaussian}. The use of a more flexible model allows for identifying complicated relationships in structural models, such as hysteretic behaviour \cite{de2021use}. The use of GPs allows skipping the selection of the order of the polynomials and offers a non-parametric fitting to the data; one may consider GPs an almost non-parametric model given that the kernel of the GP needs to be selected, and, as in every NARX-modelling case, the input and output lag numbers. GP-NARX models are also quite useful because they offer a quantification of the predictive uncertainty which is present in models built on data acquired from a physical structure \cite{worden2017uncertainty}. The uncertainty may come from inherent noise but it may also serve as a way of ensuring that the model is not extrapolating. A GP yields higher variance in the prediction when predicting with inputs away from the training domain, indicating extrapolation.

Another black-box and universal approximator which is very commonly used for system identification as a NARX model is a \textit{neural network} \cite{Bishop}. A neural-network NARX has advantages over a GP-NARX, such as the computational efficiency. Training GPs with large datasets involves high computational cost. Similarly, GPs can be multiple-input-multiple-output (MIMO) models; however such GP models require special construction to model correlations between the outputs, making most common MIMO GP models an ensemble of multiple-input-single-output models. Such an issue may miss output correlations which can be crucial for the physical system. To address such issues, multi-output neural-network NARX models have been used for identification of structural systems \cite{worden2025machine}.

The aforementioned approaches to system identification offer solutions to acquiring a model with the same (or similar) behaviour as the structure of interest. Such a model can be useful to test how the structure would behave in hypothetical scenarios. Testing the model over a range of excitations can be particularly useful when a structure exhibits nonlinear behaviour. The acquisition of a spectrum of a response or a \textit{frequency response function} (FRF) form are useful indicators of the natural frequencies of the structure and its behaviour, however, they do not fully describe the potential response of a structure. FRFs are often calculated using the H1 approximation \cite{worden2019nonlinearity}, which assumes linear behaviour. Similarly, a random excitation -- white noise in the lab or a random excitation in the field -- may not allow for exhibition of the nonlinear behaviour in a single test. Acquiring such FRFs and spectra and making a linear superposition assumption could lead to quite false predictions regarding the behaviour of the system. 

NARX models described so far are machine-learning models. They are based on assuming a parametric model and learning the values of the parameters from data. Despite their flexibility, such models come with several restrictions. As discussed already, GP and neural-network NARX models may fail to extrapolate. This inability may come from the scale difference between the training data and testing data; both types of models, in different ways, fail to make predictions outside the training range. Polynomial-NARX models may extrapolate outside the range of the training data, if they have actually captured the underlying relationship of the data. However, they may also converge to erroneous solutions because of the discussed property of polynomials having similar behaviour for a range of inputs. 

The extrapolation problem of such models becomes more evident if one considers that they refer to a single structure. NARX models use data from physical structures and provide predictions exactly for these structures; even specifically for the sensors placed on the exact positions used for the data acquisition. In contrast, mathematical modelling and formalisation of physical phenomena can be applied more widely to structures and systems of the modelled behaviour. Machine-learning approaches lack this capability, which often limits their applicability. 

To address this issue, the current work proposes the GraphNARX model based on the use of GNNs. The GNN model is one of the first machine-learning models which: (i) allow inference for different sizes of input vectors, and (ii) define learning in a local framework allowing for modular inference into larger systems. These two properties present a great alternative to fixed-size inputs, and scaling to systems which may have larger displacements which are an accumulation of many smaller local displacements. 

\subsection{Graph neural network (GNN)}
\label{sec:GNN}

The motivation for developing the GNN algorithm came from inductive biases that physical systems have which are overlooked by traditional machine-learning models -- often referred to as \textit{Euclidean} machine learning. As described in \cite{battaglia2018relational}, social networks, molecules, mechanical systems, images etc can be modelled as graphs which describe internal interactions between parts of the system. Furthermore, physical systems have a varying number of interacting elements (similarly to degrees of freedom in a structural system) whose interactions remain the same even if the size of the system changes. This property cannot be trivially captured by traditional machine-learning algorithms. A very simple example of such a system, yet quite fundamental \cite{battaglia2018relational}, is calculating the centre of mass of $N$ planets. One may train a traditional machine-learning algorithm to approximate the centre of mass using masses, and positions for the $N$ planets. However, the algorithm cannot be simply used to approximate the centre of mass of $N+1$ planets. Even worse, if one draws the masses and positions from specific distributions, the algorithm may only be able to approximate the centre if the testing masses and positions come from the same distributions without permuting them; otherwise it is called on to extrapolate and it may fail.

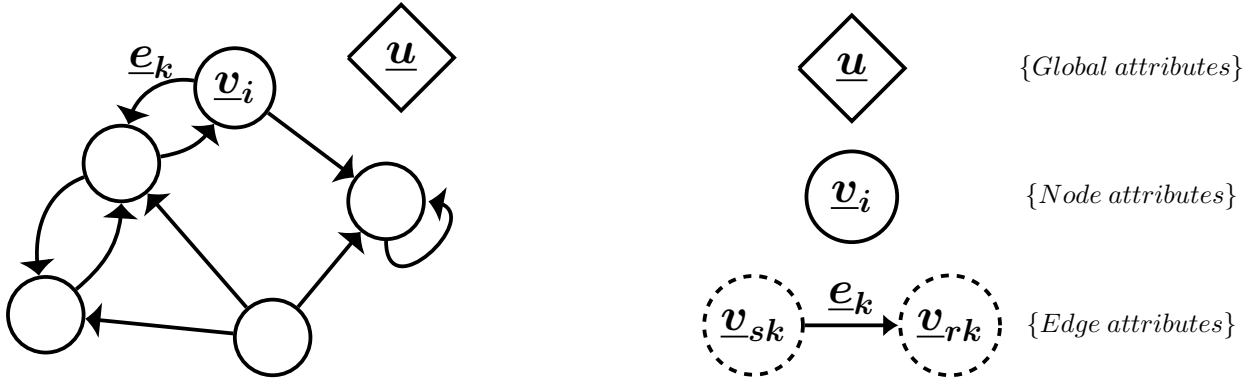
\begin{figure}
    \centering
    \begin{tikzpicture}
            
\node[circle,draw, minimum size=1cm, line width=0.5mm] (A) at  (0, 0) {};
\node[circle,draw, minimum size=1cm, line width=0.5mm] (B) at  (1, 2)  {};
\draw[-{Latex[width=5mm, length=2mm]}, line width=0.5mm] (A) to[out=40, in=270] (B);
\draw[-{Latex[width=5mm, length=2mm]}, line width=0.5mm] (B) to[out=200, in=100] (A);

\node[circle, draw, minimum size=1cm, line width=0.5mm] (C) at  (3, -0.3) {};
\node[circle, draw, minimum size=1cm, line width=0.5mm] (D) at  (2.5, 3) {\LARGE $\bm{{\underline{v}}_{i}}$};
\node[circle, draw, minimum size=1cm, line width=0.5mm] (E) at  (4.5, 1.5) {};

\draw[-{Latex[width=5mm, length=2mm]}, line width=0.5mm] (C) to (A);
\draw[-{Latex[width=5mm, length=2mm]}, line width=0.5mm] (C) to (E);
\draw[-{Latex[width=5mm, length=2mm]}, line width=0.5mm] (C) to (B);
\draw[-{Latex[width=5mm, length=2mm]}, line width=0.5mm] (D) to (E);

\draw[-{Latex[width=5mm, length=2mm]}, line width=0.5mm] (B) to[out=10, in=240] (D);
\draw[-{Latex[width=5mm, length=2mm]}, line width=0.5mm] (D) to[out=170, in=80] node[above] {\LARGE $\bm{{\underline{e}}_{k}}$} (B);

\draw[line width=0.5mm,-{Latex[width=5mm, length=2mm]},shorten >=1pt] (E) to [out=270,in=0,loop,looseness=3.4] (E);

\draw[line width=0.5mm] (4, 3.4) -- (4.7, 2.7) -- (5.4, 3.4) -- (4.7, 4.1) -- (4, 3.4);
\node[] at (4.7, 3.4) {\LARGE $\bm{{\underline{u}}}$};

\end{tikzpicture}
    \hfill
    \begin{tikzpicture}
            
\draw[line width=0.5mm] (1.8, 4) -- (2.5, 3.3) -- (3.2, 4) -- (2.5, 4.7) -- (1.8, 4);
\node[] at (2.5, 4) {\LARGE $\bm{{\underline{u}}}$};

\node[circle, draw, minimum size=1.2cm, line width=0.5mm] (D) at  (2.5, 2.3) {\LARGE $\bm{{\underline{v}}_{i}}$};

\node[circle, dashed, draw, minimum size=1cm, line width=0.5mm] (sender) at  (1.2, 0.6) {\LARGE $\bm{{\underline{v}}_{sk}}$};

\node[circle, dashed, draw, minimum size=1cm, line width=0.5mm] (receiver) at  (3.8, 0.6) {\LARGE $\bm{{\underline{v}}_{rk}}$};

\draw[-{Latex[width=3mm, length=2mm]}, line width=0.5mm] (sender) to node[above] {\LARGE $\bm{{\underline{e}}_{k}}$} (receiver);

\node[] at (6.2, 4) {\small $\{Global\ attributes\}$};
\node[] at (6.2, 2.3) {\small $\{Node\ attributes\}$};
\node[] at (6.2, 0.6) {\small $\{Edge\ attributes\}$};

\end{tikzpicture}
    \caption{General graph example.}
    \label{fig:general_graph}
\end{figure}

The GNN algorithm solves such problems by using graphs as inputs to the model and modelling the interactions of the nodes and edges of the graph. A general representation of a graph is given by,
\begin{equation}
    \pazocal{G} = (V, E)
\end{equation}
where $V$ is the set of vertices (nodes) of the graph, and $E$ is the set of edges connecting any two of the nodes. An example of a graph is given in Figure \ref{fig:general_graph}. A graph may have attributes $\bm{e}^{i}$ on its edges, on its nodes $\bm{v}^{i}$, and global attributes $\bm{u}$ of the whole graph $\pazocal{G}$. The GNN algorithm takes a graph $\pazocal{G}$ as an input and, after a series of calculations, yields an updated graph $\pazocal{G}'$ as an output. Most commonly, the topology of the updated graph is the same as the input graph; however, the features of the nodes, edges and the global ones are updated.

For the quite general case of \textit{message-passing} GNNs \cite{battaglia2018relational}, the updating is done via a series of three operations, the edge update, the node update, and the global update. The edge update yields the updated features $\bm{e}^{i\prime}$ of edges given by,
\begin{equation}
    \label{eq:edge_update_gen}
    \bm{e}^{i\prime} = \phi^{e}(\bm{e}^{i}, \bm{v}^{i,s}, \bm{v}^{i,r}, \bm{u})
\end{equation}
where $\bm{e}^{i}$ are the edge features of the $i$\textsuperscript{th} edge of the input graph, $\bm{v}^{i,s}$, and $\bm{v}^{i,r}$ are the features of the sender (tail), and receiver (head) node of the $i$\textsuperscript{th} edge, $\bm{u}$ are the global features of the input graph, and $\phi^{e}$ is a trainable parametric function, which is almost always a neural network. The second update step is the node update and is given by,
\begin{equation}
    \label{eq:node_nn}
    \bm{v}^{i\prime} = \phi^{v}(\rho^{e \to v}(E'_{i}), \bm{v}^{i}, \bm{u})
\end{equation}
where $\bm{v}^{i\prime}$ are the updated node features of the $i$\textsuperscript{th} node of the output graph, $\rho^{e \to v}(E'_{i})$ is an aggregation function applied to the updated features of the set of edges $E'_{i}$ with node $i$ as their receiver node,  $\bm{v}^{i}$ are the node features of the $i$\textsuperscript{th} node before the updating, $\bm{u}$ are the global features of the input graph, and $\phi^{v}$ is a trainable parametric function. The final update step is the global feature update step which is given by,
\begin{equation}
    \bm{u}' = \phi^{u}(\rho^{e \to u}(E'), \rho^{v \to u}(V'), \bm{u})
\end{equation}
where $\bm{u}'$ are the updated global features of the graph, $\rho^{e \to u}(E')$ is an aggregation function applied to the updated features of all edges, $\rho^{v \to u}(V')$ is an aggregation function applied to the updated features of all nodes, $\bm{u}$ are the global features of the input graph, and $\phi^{u}$ is also a trainable parametric function. 

According to the formulation of the above GNN model, the updating from a graph $\pazocal{G}$ to a graph $\pazocal{G}'$ is done via three functions (most often neural networks) $\phi^{e}, \phi^{v},$ and $\phi^{u}$. One of the main novelties of the GNN algorithm, which allows for inference in graphs of different sizes, is the use of the aggregation functions $\rho$. Common selections for the aggregation functions are averaging, summation, maximum, and minimum. According to the requirements of the problem, a subset of the update functions may be used. Moreover, considering that a tuple $(\phi^{e}, \phi^{v}, \phi^{u})$ comprises one GNN layer, several layers can be applied for the update from $\pazocal{G}$ to $\pazocal{G}'$. Such a selection may be done because of the physics of the problem, e.g.\ because of the need to propagate information between nodes which do not belong in the immediate neighbourhood of each other. For more information, the interested reader may refer to \cite{battaglia2018relational}.

GNNs have been used for solving different types of problems. For example, the connectivity between a set of objects can be set as an objective as an edge-classification problem, where the target values are the labels of the edges of the updated graph \cite{kipf2016variational}. Similarly, node classification can be performed by considering the class of the nodes as the target \cite{Kipf}. The global features are used to classify whole graphs, e.g.\ in \cite{wu2018moleculenet}, a GNN is used for the classification of molecules. For structural dynamics, GNNs have been used for various tasks. In \cite{tsialiamanis2021foundations}, graphs are used to represent structures and to predict their natural frequency. In \cite{PBSHM5MSSP}, GNNs are used to classify structures according to their similarity with a view to maximising the probability of positive transfer between similar structures. In \cite{duthe2023graph}, GNNs are used to predict the flow around air-foils with sparse sensing information. In \cite{haywood2026piggo}, a method is presented to model the dynamics of structures in the state space, using known linear dynamics and learning the residual unknown physics. The approach is also used in combination with a Kalman filter for virtual sensing. The current work follows a different approach in identifying structures via the NARX formulation and considering lagged inputs to the model. Such an approach does not require the knowledge of any physics and presents a simpler approach to virtual sensing. 

\subsection{GNN as a system identification model}
\label{sec:gnn_sid}

Modelling structures as discrete systems is a widely-used approach in engineering. From quite simple lumped-mass systems to the FE method, discretisation allows for modelling structural dynamics and identifying structures. Such a widely-used approach makes the use of graphs a natural choice, especially when it comes to machine learning. Using a graph which encodes the geometry of a structure, the nodes may represent the degrees of freedom, and the edges may represent the connectivity of these degrees of freedom, i.e.\ the neighbouring nodes. In a continuous structure, the existence of an edge between two nodes represents the existence of material between the two, not just spatial proximity. The features of the nodes may comprise the state of the system, or, in a NARX setting, the lagged displacements (or velocities or accelerations). Moreover, the node attributes may include the input forces to that point of the structure, and the mass attributed to that point.

\begin{figure}
    \centering
    \includegraphics[width=\linewidth]{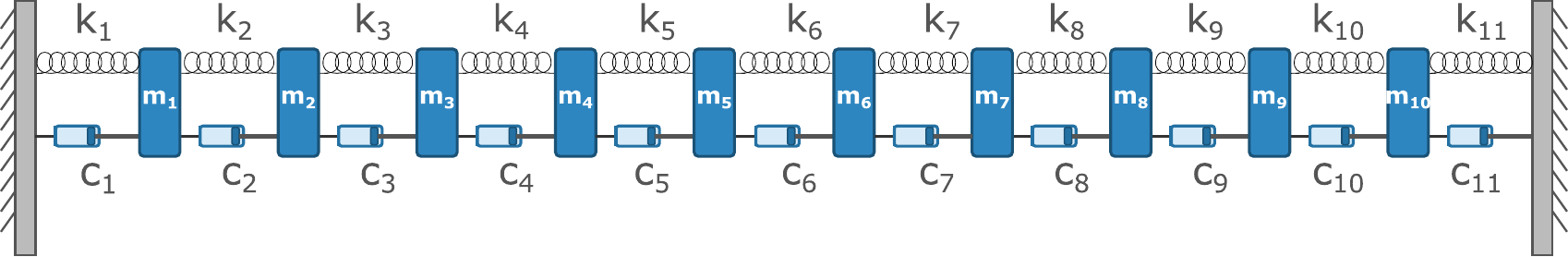}
    \caption{Example mass-spring system of ten degrees of freedom.}
    \label{fig:10_dof}
\end{figure}

\begin{figure}
    \centering
    \includegraphics[width=\linewidth]{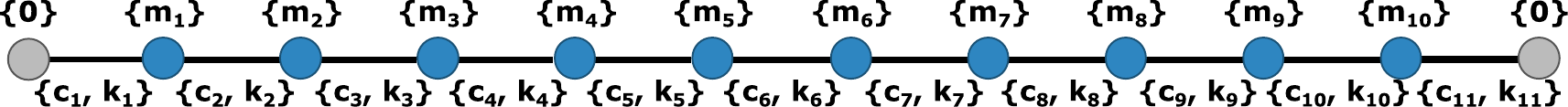}
    \caption{Graph representation of the mass-spring system of Figure \ref{fig:10_dof}.}
    \label{fig:10_dof_graph_simple}
\end{figure}

A discretised structure can be straightforwardly transformed into a graph. A simple transformation of the mass-spring system of Figure \ref{fig:10_dof} is shown in Figure \ref{fig:10_dof_graph_simple}. A simple approach to apply the NARX formulation using a graph is to use the concatenated delayed displacements and input forces as the node features. Therefore, the node features of the $i$\textsuperscript{th} node are given by,
\begin{equation}
    \bm{v}^{i}_{t} = \{y^{i}_{t}, y^{i}_{t-1}..., y^{i}_{t-n_{y}}, x^{i}_{t}, x^{i}_{t-1}..., x^{i}_{t-n_x}\}
    \label{eq:node_feats}
\end{equation}
where $y_{t}^{i}$ is the displacement of the $i$\textsuperscript{th} node at time $t$, and $x^{i}_{t}$ is the force applied to the same node at time $t$. If all nodes of the system have the same characteristics, no further features are required to be encoded in $\bm{v}^{i}_{t}$, otherwise, such features need to be concatenated with the rest of the node-feature vectors. For example, if the nodes have different masses, the masses need to be encoded in the $\bm{v}$ vector. Similarly, the edges encode the connection between the nodes. Such connections refer to segments of structural members, therefore the features of the edges of the graph must include information regarding the structural characteristics of such segments. In the simplified example of a lumped-mass system, the edges may have features given by,
\begin{equation}
    \bm{e}^{i} = \{k^{i}, c^{i}\}
    \label{eq:edge_feats}
\end{equation}
where $k^{i}$ is the segment's stiffness, and $c^{i}$ is the segment's damping coefficient. 

Following the above formalisation, the GNN model can be trained in order to predict the displacement at $t+1$ of every node, i.e.\ $\bm{y}_{t+1} = f(\pazocal{G}_{t})$, where $f$ is the GNN model and $\pazocal{G}$ is a graph whose edge and node features are given by equations (\ref{eq:node_feats}) and (\ref{eq:edge_feats}). Training such a model is quite similar to the training of traditional neural-network NARX models via backpropagation. The model is trained with the nodal displacements at $t+1$ as the target.

The aim of the current work is to build a GraphNARX model which can generalise the system identification to structures of higher number of degrees of freedom. Because the GNN model uses neural networks for the predictions, a first challenge in order to achieve generalisation in different degrees of freedom is the scale of the displacements of the target structure. Neural networks are great approximators; however, only within the range of their training domain, and, quite often only within the $[-1, 1]$ range (which is imposed by their activation functions). A second challenge which is to be addressed by the current work is that the material properties may not be known. Thus, the definition of the edge features is not trivial. 

To address these challenges, modifications are made to the straightforward approach described so far. A first modification is that the edge features do not contain the material characteristics. The identified structure is assumed to be made of the same material with unknown characteristics. In case of segments of different dimensions between the nodes, one could use the features to encode the variation in the size of the segments; however, in the current work, segments of the same characteristics are considered. Guided by the physics of the problem, the edges should describe how the forces are transmitted from one node to the other, thus their features $\bm{e}_{t}^{i}$ are given by,
\begin{equation}
    \bm{e}_{t}^{i} = \{y^{i, r}_{t} - y^{i, s}_{t}, y^{i, r}_{t-1} - y^{i, s}_{t-1}..., y^{i, r}_{t-n_{y}} - y^{i, s}_{t-n_{y}}\}
\end{equation}
where $y^{i, r}_{t}$ and $y^{i, s}_{t}$ are the displacements of the receiver (head) and sender(tail) nodes of the $i$\textsuperscript{th} edge respectively. Such an approach is motivated by the fact that the forces of the edges are described by the relative displacements of neighbouring nodes, rather than absolute displacements.

\begin{figure}
    \centering
    \includegraphics[width=0.85\linewidth]{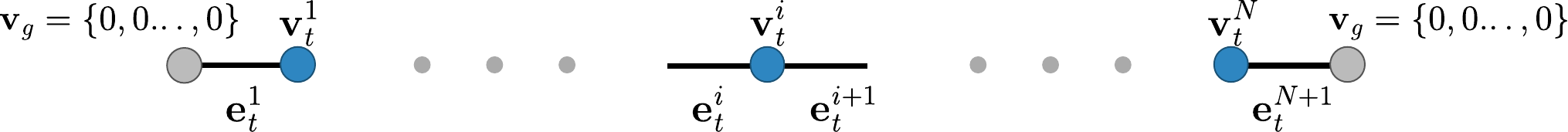}
    \caption{Input graph representation of mass-spring system for the proposed GraphNARX model.}
    \label{fig:graphNARX}
\end{figure}

The input graph to the GraphNARX model is schematically shown in Figure \ref{fig:graphNARX}. The node features, as shown in equation (\ref{eq:node_feats}), can be used without any modification, given that the nodes are considered identical. In a mass-spring system this assumption means that all nodes have the same mass. If masses were different, the mass should also be included in the node features $\bm{v}^{i}_{t}$ to allow the GNN to learn the effects of inertia in the dynamics of the system. In the current work, no global characteristics are considered for the systems, however, if an environmental parameter varied between the systems or in time, that could be encoded in the global feature vector $\bm{u}_{t}$; such an approach is presented in \cite{tsialiamanis2021foundations}, where GNNs are used to estimate the normal-condition natural frequencies of structures under temperature variations. 

The GraphNARX update step is defined by two models. The first is the edge-update model $\phi^{e}$, which yields,
\begin{equation}
    \bm{e}^{i\prime} = \phi^{e}(\bm{e}^{i}) = \phi^{e}(y^{i, r}_{t} - y^{i, s}_{t}, y^{i, r}_{t-1} - y^{i, s}_{t-1}..., y^{i, r}_{t-n_{y}} - y^{i, s}_{t-n_{y}})
    \label{eq:edge_update}
\end{equation}
and the node-update model $\phi^{v}$, which yields,
\begin{equation}
    \label{eq:node_update}
    \Delta \hat{y}^{i}_{t+1} = \phi^{v}(\rho^{e \to v}(E'_{i}), \bm{v}^{i}) = \phi^{v}(\sum_{j \in E'_{i}}\bm{e}^{j\prime}, y^{i}_{t}, y^{i}_{t-1}..., y^{i}_{t-n_{y}}, x^{i}_{t}, x^{i}_{t-1}..., x^{i}_{t-n_x})
\end{equation}
where $\Delta \hat{y}^{i}_{t+1}$ is the estimate of the displacement increment at the $i$\textsuperscript{th} node at time $t+1$, and $E'_{i}$ is the set of edges with $i$ as their receiver node. Modelling the increment allows for the total predicted displacement to be equal to,
\begin{equation}
    \hat{y}^{i}_{t+1} = y^{i}_{t} + \Delta \hat{y}^{i}_{t+1}
\end{equation}
Such an approach is useful when using neural networks, because the increment is more commonly within the training domain of the neural network, even for larger structures. For the ground nodes, the displacements are imposed to be equal to zero. The model is trained with a mean-squared error (MSE) loss function given by,
\begin{equation}
    \label{eq:loss}
    \pazocal{L} = \frac{1}{M}\sum_{j=1}^{M}\frac{1}{N^{j}}\sum_{i=1}^{N^{j}}(\hat{y}^{i, j}_{t+1} - {y}^{i, j}_{t+1})^{2}
\end{equation}
where $M$ is the total number of training samples, $N^{j}$ is the number of nodes in the $j$\textsuperscript{th} sample, and $\hat{y}^{i, j}_{t+1}$ are the model predictions of the displacement at the next timestep of the $j$\textsuperscript{th} training sample of the $i$\textsuperscript{th} node of the graph and ${y}^{i, j}_{t+1}$ the actual displacements.

\section{System identification application}
\label{sec:sid_app}

\subsection{A ten-degree-of-freedom system}
\label{sec:10dof}

As mentioned, the GraphNARX system is proposed as a system-identification model. More specifically, it is used to identify a model which predicts the displacements of nodes of a system given the inductive bias of the connectivity of the nodes, i.e.\ the geometry of the structure. Traditional NARX models allow for interactions between all degrees-of-freedom of the system in order to predict the next-step displacements. The proposed GraphNARX method only allows predictions of displacements of every node based on the neighbouring nodal displacements. Such an approach is arguably more natural and is what may allow extrapolation to different geometries.

A simple mass-spring system was considered to evaluate the efficiency of the method. The mass-spring system of ten degrees of freedom is shown in Figure \ref{fig:10_dof}. The springs and dampers system was considered to have the same value of parameters. The values of the parameters are mass equal to $m=1$kg, $c=1.0$Ns/m, and $k=1000$N/m. To simulate the system, a fourth-order Runge-Kutta algorithm was used with a $0.01$s timestep and three datasets were generated, one for training, one for model selection (validation) and one for testing the selected model. Each dataset comprises $20000$ timesteps. The results presented here are exclusively for the testing dataset; i.e.\ unseen data for the model. For regularisation, the data were corrupted with $1\%$ white noise; however, the results are presented comparing the predictions with the noise-free data, in order to test the model's ability to perform such a denoising process. % IDEA: corrupt the data with noise in the Multi-step prediction

The loss function for training is the one given by equation (\ref{eq:loss}), i.e.\ the \textit{one-step-ahead} (OSA) loss function. Initially, training was performed using Adam \cite{kingma2014adam} with a learning rate of $10^{-3}$ for $100$ epochs. Such a learning rate resulted in low OSA errors; however, when the \textit{model predictive error} (MPE) (i.e.\ recursively feeding the model's own previous predictions forward over a multi-step prediction horizon) was evaluated, it was observed that the error was significantly higher. In order to fine-tune the model further, training was performed using two stages, $100$ epochs with a learning rate of $10^{-3}$ and another $100$ epochs with a lower learning rate of $10^{-4}$. This approach improved the MPE results significantly. The errors are calculated using the \textit{normalised mean-square error} (NMSE) which is given by,
\begin{equation}
    \label{eq:NMSE}
    NMSE = \frac{100}{M}\sum_{j=1}^{M}\frac{1}{N^{j}}\sum_{i=1}^{N^{j}}(\hat{y}^{i, j}_{t+1} - {y}^{i, j}_{t+1})^{2} / \sigma_{y}^{2}
\end{equation}
where $M$ is the total number of training samples, $N^{j}$ is the number of nodes in the $j$\textsuperscript{th} sample, and $\hat{y}^{i, j}_{t+1}$ are the model predictions of the displacement at the next timestep of the $j$\textsuperscript{th} training sample of the $i$\textsuperscript{th} node of the graph, ${y}^{i, j}_{t+1}$ the actual displacements, and $\sigma_{y}^{2}$ is the variance of the displacements of the system. The NMSE is more representative of the fit of the model than the MSE because it compares the error with the variance of the data. From the authors' experience, NMSE values lower than $1\%$ show an excellent fit to the data, and values below $5\%$ indicate a well-fitted model. From its definition, NMSE values close to $100\%$ mean that the model only predicts the mean value of the data, potentially meaning that the required information to perform inference is not in the input variables.

\begin{table}[h!]
    \caption{GNN network sizes considered in training}
    \label{tab:net_sizes}
    \begin{center}
        \begin{tabular}{|c|c|c|}
            \hline
            \textbf{ID} & $\phi^{e}$ & $\phi^{v}$ \\
            \hline
            \textbf{GNN 1}   &   $(n_{y}, 10, 5)$  &  $(5+n_{y}+n_{x}, 15, 1)$ \\
            \hline
            \textbf{GNN 2}     &   $(n_{y}, 15, 7)$  &  $(7+n_{y}+n_{x}, 20, 1)$ \\
            \hline
            \textbf{GNN 3}     &   $(n_{y}, 20, 10)$  &  $(10+n_{y}+n_{x}, 25, 1)$ \\
            \hline
        \end{tabular}
    \end{center}
\end{table}

The GraphNARX model is defined by a set of hyperparameters which need optimisation. The first set of hyperparameters is the architecture of the edge-update model $\phi^{e}$ and node-update model $\phi^{v}$ from equations (\ref{eq:edge_update}) and (\ref{eq:node_update}) correspondingly. The second set of hyperparameters, as in every type of NARX models, is the lags. For the architecture of the neural networks of the GNN model, three sets of hidden layers were considered. The architectures are shown in Table \ref{tab:net_sizes}. For each one of them, the validation OSA error was used for model selection and the displacement and force lags were selected by considering the values after which the error did not drop significantly. Following this approach, the architecture of GNN 2 with lag values of $6$ for the displacement lag $n_{y}$ and $2$ for the force lag $n_{x}$ was used (the initial lag values considered were $2$ for $n_{y}$ and $1$ for $n_{x}$).

\begin{figure}
    \centering
    \includegraphics[width=0.7\linewidth]{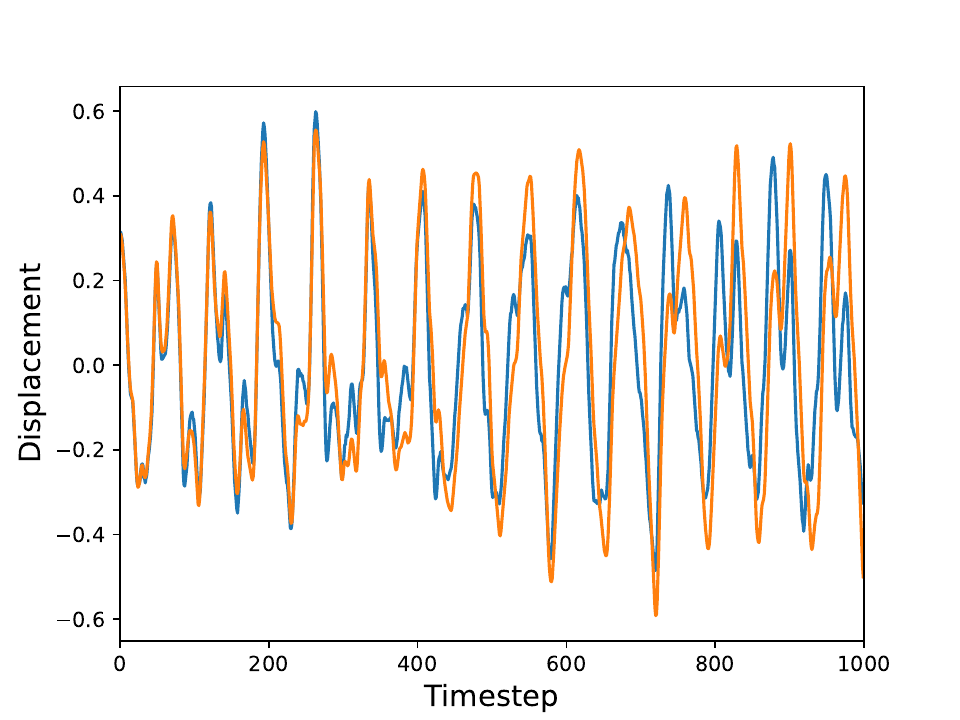}
    \caption{Example of actual displacements (blue curve) compared to multi-step-ahead predictions of the GraphNARX model (orange curve) for the sixth degree-of-freedom of the ten-degree-of-freedom system.}
    \label{fig:example_same_size_dof_6}
\end{figure}

\begin{figure}
    \centering
    \includegraphics[width=0.7\linewidth]{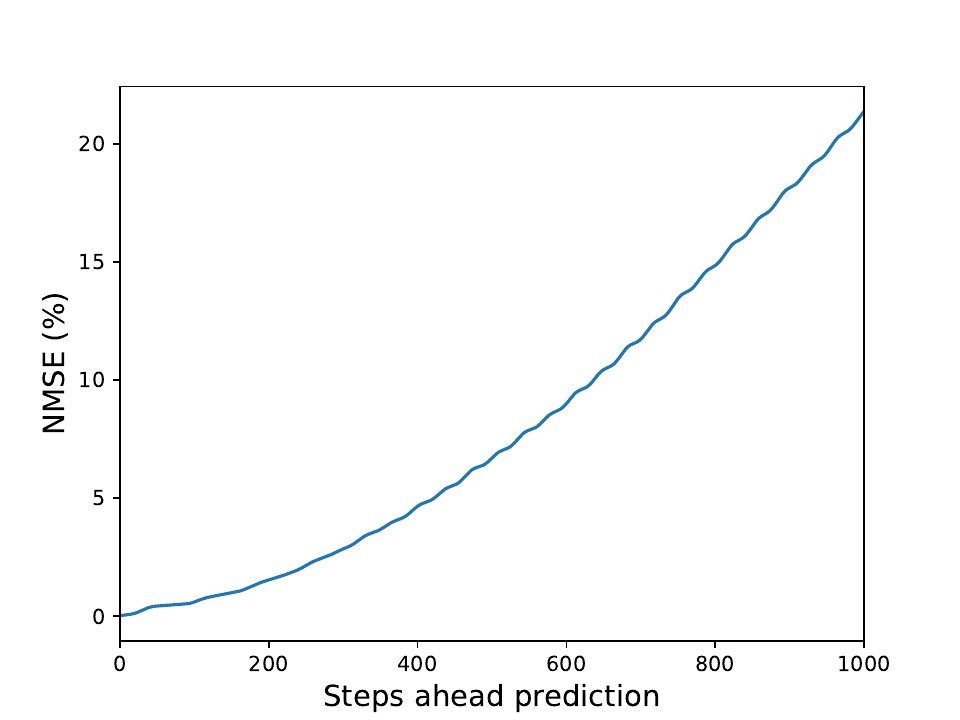}
    \caption{Average NMSE for different values of timestep-ahead predictions of the GraphNARX model for the testing dataset of the ten-degree-of-freedom system.}
    \label{fig:NMSE_same_size}
\end{figure}

The testing-data OSA error for the selected model was $0.0204\%$, which is significantly low and also shows that the model has learnt to perform denoising. Using this model, multi-step-ahead predictions were made for the testing dataset for $1000$ timesteps. An example of the predictions compared to the actual displacements of the sixth degree of freedom of the system is shown in Figure \ref{fig:example_same_size_dof_6}. The average NMSE for the testing dataset for different values of steps ahead is shown in Figure \ref{fig:NMSE_same_size}. In the figure, it can be seen that the error increases for longer-horizon predictions. For $400$ timesteps ahead the average NMSE is lower than $5\%$ which indicates a good fit for such a horizon. It is also worth noting that the error increases almost linearly after $400$ steps. 

\subsection{Extrapolation to a 20-degree-of-freedom system}
\label{sec:20_dof}

The same GraphNARX model trained using the ten-degree-of-freedom system dataset was tested on a 20-degree-of-freedom dataset. The larger system was considered with the same values for mass, damping, and stiffness parameters (i.e.\ a larger structure made of the same material as the ten-degree-of-freedom structure). The same simulation parameters were used and a $20000$-timestep dataset was created. Only one dataset is generated for the larger system, since there is no training or fine-tuning involved in this case study, only testing the already-trained model. The trained model was used for OSA predictions and for multi-step-ahead predictions. The OSA NMSE was equal to $0.02283\%$, which is also quite low and on the same order of magnitude as the OSA NMSE of the ten-degree-of-freedom structure. 

\begin{figure}
    \centering
    \includegraphics[width=0.7\linewidth]{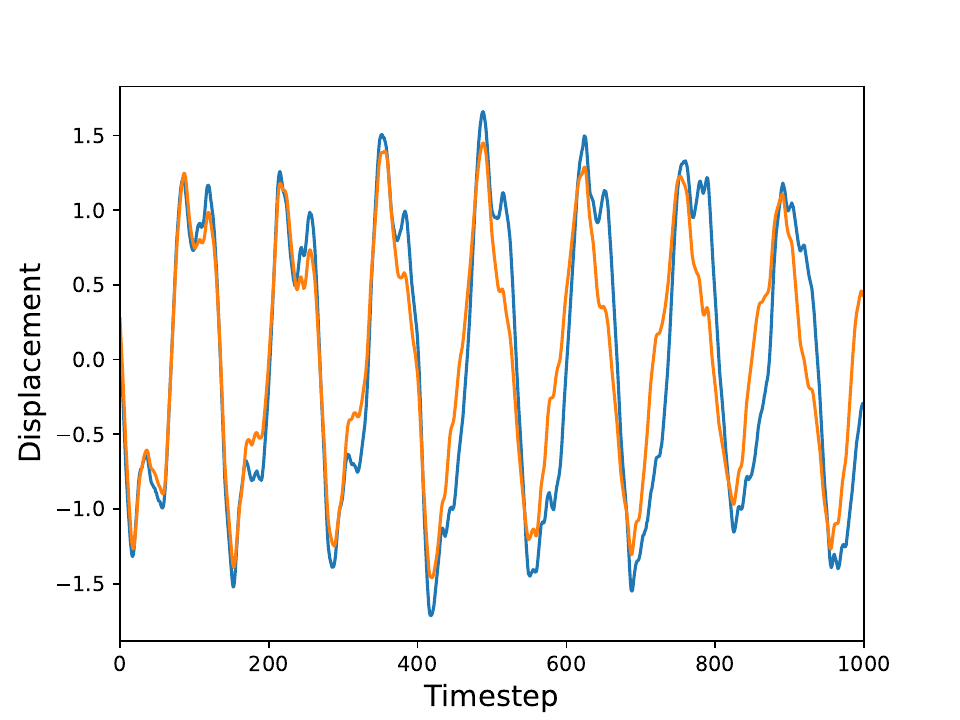}
    \caption{Example of actual displacements (blue curve) compared to multi-step-ahead predictions of the GraphNARX model (orange curve) for the tenth degree-of-freedom of the 20-degree-of-freedom system.}
    \label{fig:example_larger_size_dof_10}
\end{figure}

\begin{figure}
    \centering
    \includegraphics[width=0.7\linewidth]{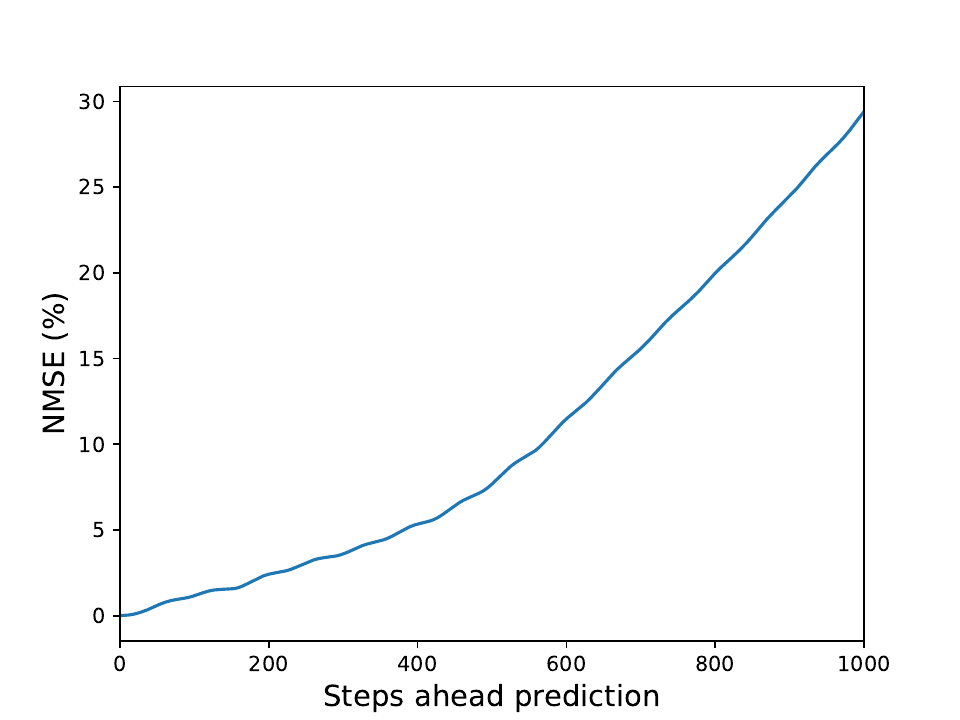}
    \caption{Average NMSE for different values of timestep-ahead predictions of the GraphNARX model for the testing dataset of the 20-degree-of-freedom system.}
    \label{fig:NMSE_larger}
\end{figure}

For multi-step ahead predictions, the same set-up as with the ten-degree-of-freedom system was used. The error was evaluated for different numbers of steps ahead and the average NMSE is calculated. An example of the actual displacements compared to the predictions of the model for $1000$ steps ahead for the tenth degree of freedom of the larger system is shown in Figure \ref{fig:example_larger_size_dof_10}. Similarly to the ten-degree-of-freedom system, the predictions are quite close to the actual displacements at the beginning and, later, larger discrepancies are observed. This decrease in accuracy is also evident in Figure \ref{fig:NMSE_larger}, where the average NMSE over the larger-structure testing dataset is shown. A similar behaviour of an NMSE lower than $5\%$ for 400-step-ahead predictions is observed, however the error increases (almost linearly) and with higher rate than the ten-degree-of-freedom system. This faster increase might be because of the accumulation of more errors because of the larger number of degrees of freedom.

\section{GraphNARX for virtual sensing}
\label{sec:VS_graphnarx}

A special feature of the GraphNARX model which distinguishes it from traditional NARX models is that the information is transferred only between neighbouring nodes. Moreover, the information is transferred using exactly the same way (same neural network model) from one node to the other. This feature of the GraphNARX model, and GNN-based models in general, can be particularly useful for \textit{virtual sensing} \cite{haywood2026piggo, avendano2021virtual}. A very simple way to use the GraphNARX model for virtual sensing is described here. To perform virtual sensing using the proposed model, two small modifications to the algorithm of Section \ref{sec:graphnarx} are performed. 

The first modification is that the inputs to the nodes which are assumed to not have a sensor are masked. This masking simply means that all past displacements and forces to these nodes are considered to be equal to zero. Randomly masking nodal inputs during training forces the model to learn how to infer the next-step displacements using partial data. It is worth noting here that, since the GraphNARX model infers the next-step displacements and the nodal forces of the masked nodes are missing, the errors are expected to be significantly higher than in the previous case study. The second modification is, naturally, in the loss function of equation (\ref{eq:loss}). The aforementioned equation considers the loss at every node of the network. In the absence of sensors on some points of the structure, the loss may only be calculated on the available sensors. Thus, the loss function is modified and is given by,
\begin{equation}
    \pazocal{L}_{VS} = \frac{1}{M}\sum_{j=1}^{M}\frac{1}{N^{j}}\sum_{i=1}^{N^{j}}m_{i}(\hat{y}^{i, j}_{t+1} - {y}^{i, j}_{t+1})^{2}, \quad m_{i}\in\{0, 1\}
\end{equation}
where the extra parameter $m_{i}$ is equal to $1$ if data from the $i$\textsuperscript{th} sensor are available, and $0$ if not. Training is performed by considering a specific masking vector $\bm{m}=[m_{1}, m_{2}..., m_{N}]$ for the $N$ sensor points. In the current work, a more general framework is considered where the model is trained with a masking vector sampled randomly during training with a masking rate of $50\%$, i.e.\ $m_{i} \sim Bern(0.5)$, where $Bern$ is the binary Bernoulli distribution.

The same training and hyperparameter-optimisation process was followed as in the previous case-study. This time, the selected lag values were significantly higher than before, namely, $n_{y}=17$, and $n_{x}=5$. The selected GNN architecture was GNN 3 from Table \ref{tab:net_sizes}. 

\begin{figure}
    \centering
    \includegraphics[width=0.7\linewidth]{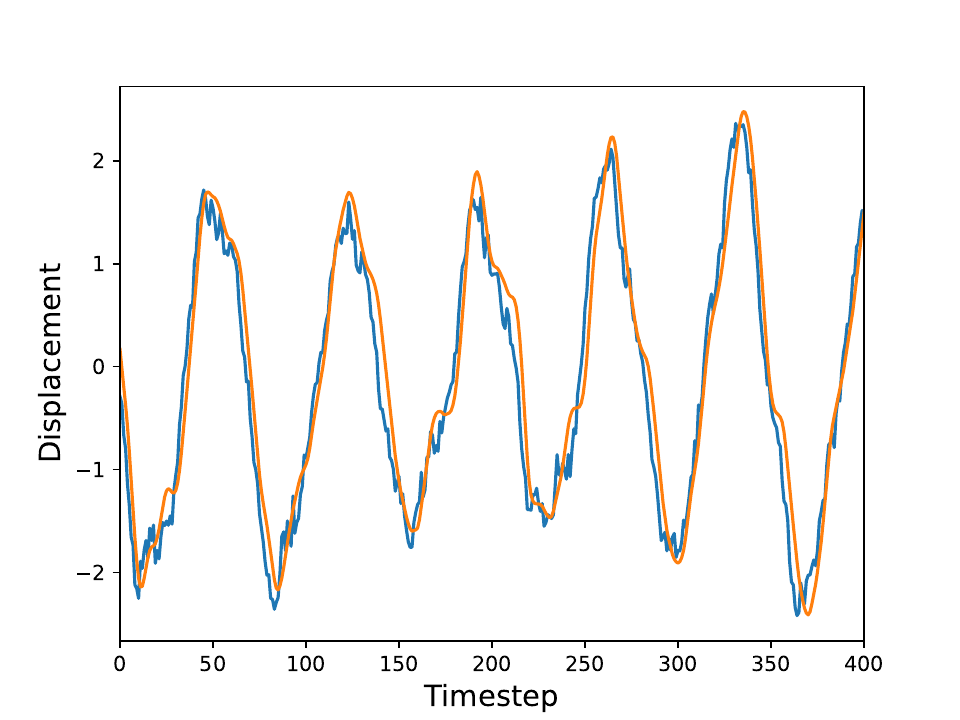}
    \caption{Example of actual displacements (blue curve) and predicted displacements (orange curve) for the virtual-sensing application for the sixth node of the ten-degree-of-freedom system, which was assumed to not have a sensor. The mask for the system was $\bm{m}=[1, 0, 1, 0, 1, 0, 1, 0, 1, 0]$, where $1$ indicates an available sensor and $0$ a missing sensor}
    \label{fig:VS_example_1_same_size}
\end{figure}

To test the accuracy of the model, two sensor set-ups are considered. The first is $\bm{m}=[1, 0, 1, 0, 1, 0, 1, 0, 1, 0]$. In this set-up, only one sensor is unavailable between two available ones. The ground-node displacements are assumed to be known and equal to zero. In this case, the NMSE over the testing dataset of the ten-degree-of-freedom system was equal to $4.994255\%$. An example of the actual versus the predicted displacements for the sixth node is shown in Figure \ref{fig:VS_example_1_same_size}. The second sensor set-up is $\bm{m}=[0, 0, 1, 0, 0, 1, 0, 0, 1, 1]$. The latter set-up has $60\%$ missing sensors and includes cases of two missing sensors between two available sensors. The NMSE in this case was equal to $6.35518\%$, which is higher than before. The increase might be attributed to the higher percentage of missing sensors and the transfer of information further between nodes of known displacements. Nevertheless, the error is not significantly high.

\begin{figure}
    \centering
    \includegraphics[width=0.7\linewidth]{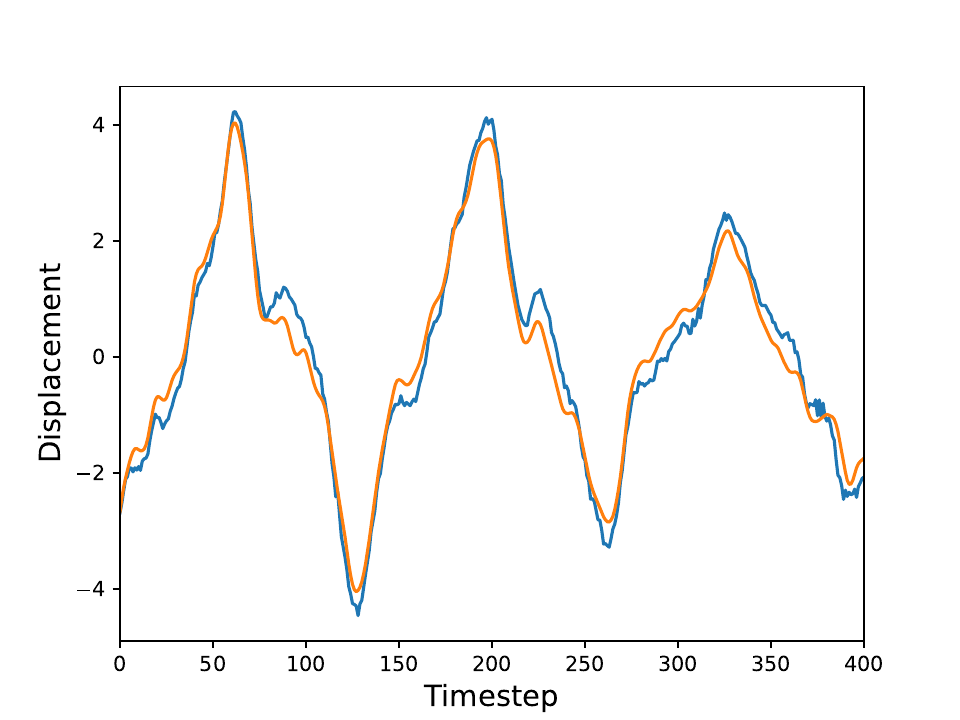}
    \caption{Example of actual displacements (blue curve) and predicted displacements (orange curve) for the virtual-sensing application for the 11\textsuperscript{th} node of the 20-degree-of-freedom system, which was assumed to not have a sensor. The mask for the system was $\bm{m}=[0, 0, 1, 0, 0, 1...0, 0, 1]$, where $1$ indicates an available sensor and $0$ a missing sensor}
    \label{fig:VS_example_2_larger}
\end{figure}

The same model was tested for its virtual-sensing capabilities on a 20-degree-of-freedom system, the same as the one used in the previous section. Similarly to the ten-degree-of-freedom system, two masks were used for the evaluation, the first was $\bm{m}=[0, 1, 0, 1..., 0, 1]$, i.e.\ one unavailable sensor between two available ones. In this case the NMSE for the testing data was equal to $1.925\%$, which is lower than in the case of the ten-degree-of-freedom system. Similarly, the second mask was $\bm{m}=[0, 0, 1, 0, 0, 1..., 0, 0, 1]$, i.e.\ two unavailable sensors between two available ones. As expected, the error in this case was higher than the case of higher number of available sensors, namely, the NMSE was equal to $2.505196\%$, lower than that of the ten-degree-of-freedom system. An example of actual displacements compared to predicted ones for a point of a missing sensor is given in Figure \ref{fig:VS_example_2_larger}.

\section{Discussion and conclusions}
\label{sec:discussion}

The current work introduces a graph-based NARX model for structural dynamics, the GraphNARX model. The approach is motivated by the capability of GNNs to learn local relationships between elements of the graph and to generalise to larger graphs. Similarly, the aim of the current work is to define a NARX model which learns the behaviour of structures of a certain number of degrees of freedom and may generalise to structures made of the same material however of larger dimensions and higher number of degrees of freedom. Because of the way that GNNs learn local relationships, the proposed model is also used for learning with data from a subset of the degrees of freedom of the structures and inferring the missing data, i.e.\ virtual sensing. 

The GraphNARX model was tested on a simple simulated mass-spring structure assuming the same properties for all springs. The one-step-ahead error was used for training the model (as in most NARX models) and the predictive accuracy was tested for multiple-step-ahead recurrent predictions. The model was trained on a ten-degree-of-freedom system. It achieved a very low NMSE ($0.0204\%$) for one-step-ahead predictions and a relatively low NMSE for multi-step-ahead predictions; however, the latter was almost linearly increasing with the prediction horizon. The same model was then used for modelling a 20-degree-of-freedom system assumed to have the same structural-parameter values; this application could be the case of training on a beam of certain length and testing on a beam of double the length. The one-step-ahead error for the larger system was also very low ($0.02283\%$), and the multi-step-ahead prediction had a similar behaviour, in increasing almost linearly in time.

The GraphNARX model was also trained on data from the ten-degree-of-freedom structure assuming that sensors were missing. The model was then evaluated on the accuracy of the next-step prediction for the positions of the missing sensors. The errors were again quite low (around $5\%$) for the ten-degree-of-freedom system. The generalisation capability of the model was again tested on a 20-degree-of-freedom system. The error in this case was even lower (around $2\%$). 

Overall the results are quite encouraging. The GraphNARX model appears to have the capability of training on structures of a certain size and generalising to larger structures. More extensive fine-tuning is certainly required. The evaluated GNN architectures were quite limited in this study. Moreover, the model-selection criterion in this case was limited to the one-step-ahead error in the validation dataset. A more complex criterion considering multi-step-ahead predictions could also be taken into account. The loss function may also be extended to include multi-step-ahead predictions similar to a recurrent neural network \cite{rumelhart1985learning}. Naturally, evaluation of the model in experimental structures is needed.

%%%%%%%%%%%%%%%%%%%%
% Acknowledgements %
%%%%%%%%%%%%%%%%%%%%

\section*{Acknowledgements}
K.W. gratefully acknowledges the support of the Engineering and Physical Sciences Research Council (EPSRC) Open Fellowship via Grant reference EP/X040852/1. For the purpose of open access, the authors have applied a Creative Commons Attribution (CC BY) licence to any Author Accepted Manuscript version arising.

%%%%%%%%%%%%%%
% References %
%%%%%%%%%%%%%%

% Include your references below, e.g. by appending a '.bib' file. 
% The style, IEEE, is already pre-defined in the class description.

\bibliography{References}

@book{bathe2006finite,
  title={Finite {E}lement {P}rocedures},
  author={Bathe, K.-J.},
  year={2006},
  publisher={Klaus-Jurgen Bathe}
}

@article{noel2017nonlinear,
  title={Nonlinear system identification in structural dynamics: 10 more years of progress},
  author={No{\"e}l, J.-P. and Kerschen, G.},
  journal={{M}echanical {S}ystems and {S}ignal {P}rocessing},
  volume={83},
  pages={2--35},
  year={2017},
  publisher={Elsevier}
}

@article{kiran2025state,
  title={A state-of-the-art review of Bayesian finite element model updating techniques for structural systems},
  author={Kiran, R.P. and Das, A. and Bansal, S.},
  journal={Probabilistic {E}ngineering {M}echanics},
  volume={80},
  pages={103761},
  year={2025},
  publisher={Elsevier}
}

@article{liu2021machine,
  title={Machine-learning-based methods for output-only structural modal identification},
  author={Liu, D. and Tang, Z. and Bao, Y. and Li, H.},
  journal={Structural {C}ontrol and {H}ealth {M}onitoring},
  volume={28},
  number={12},
  pages={e2843},
  year={2021},
  publisher={Wiley {O}nline {L}ibrary}
}

@inproceedings{rogers2017grey,
  title={On a grey box modelling framework for nonlinear system identification},
  author={Rogers, T.J. and Holmes, G.R. and Cross, E.J. and Worden, K.},
  booktitle={{S}pecial {T}opics in {S}tructural {D}ynamics, Volume 6: {P}roceedings of the 35th {IMAC}, {A} {C}onference and {E}xposition on {S}tructural {D}ynamics 2017},
  pages={167--178},
  year={2017},
  organization={Springer}
}

@inproceedings{wigren2013three,
  title={Three free data sets for development and benchmarking in nonlinear system identification},
  author={Wigren, T. and Schoukens, J.},
  booktitle={2013 European {C}ontrol {C}onference ({ECC})},
  pages={2933--2938},
  year={2013},
  organization={IEEE}
}

@article{battaglia2018relational,
  title={Relational inductive biases, deep learning, and graph networks},
  author={Battaglia, P.W. and Hamrick, J.B. and Bapst, V. and Sanchez-Gonzalez, A. and Zambaldi, V. and Malinowski, M. and Tacchetti, A. and Raposo, D. and Santoro, A. and Faulkner, R. and others},
  journal={arXiv preprint arXiv:1806.01261},
  year={2018}
}

@book{billings2013nonlinear,
  title={Nonlinear system identification: {NARMAX} methods in the time, frequency, and spatio-temporal domains},
  author={Billings, S.A.},
  year={2013},
  publisher={John Wiley \& Sons}
}

@article{kopsaftopoulos2012vector,
  title={Vector-dependent functionally pooled {ARX} models for the identification of systems under multiple operating conditions},
  author={Kopsaftopoulos, F.P. and Fassois, S.D.},
  journal={{IFAC} {P}roceedings {V}olumes},
  volume={45},
  number={16},
  pages={310--315},
  year={2012},
  publisher={Elsevier}
}

@techreport{rumelhart1985learning,
  title={Learning internal representations by error propagation},
  author={Rumelhart, D.E. and Hinton, G.E. and Williams, R.J.},
  year={1985}
}

@book{worden2019nonlinearity,
  title={Nonlinearity in structural dynamics: detection, identification and modelling},
  author={Worden, K. and Tomlinson, G.R.},
  year={2019},
  publisher={CRC Press}
}

@article{kerschen2006past,
  title={Past, present and future of nonlinear system identification in structural dynamics},
  author={Kerschen, G. and Worden, K. and Vakakis, A.F. and Golinval, J.-C.},
  journal={Mechanical {S}ystems and {S}ignal {P}rocessing},
  volume={20},
  number={3},
  pages={505--592},
  year={2006},
  publisher={Elsevier}
}

@incollection{worden2014gaussian,
  title={On Gaussian process {NARX} models and their higher-order frequency response functions},
  author={Worden, K. and Manson, G. and Cross, E.J.},
  booktitle={Solving {C}omputationally {E}xpensive {E}ngineering {P}roblems: {M}ethods and {A}pplications},
  pages={315--335},
  year={2014},
  publisher={Springer}
}

@book{williams2006gaussian,
  title={Gaussian processes for machine learning},
  author={Williams, C.K.I. and Rasmussen, C.E.},
  volume={2},
  number={3},
  year={2006},
  publisher={MIT press Cambridge, MA}
}

@article{de2021use,
  title={On the use of the {GP-NARX} model for predicting hysteresis effects of bolted joint structures},
  author={de Oliveira Teloli, R. and Villani, L.G.G. and da Silva, S. and Todd, M.D.},
  journal={Mechanical {S}ystems and {S}ignal {P}rocessing},
  volume={159},
  pages={107751},
  year={2021},
  publisher={Elsevier}
}

@article{worden2017uncertainty,
  title={Uncertainty bounds on higher-order {FRFs} from {G}aussian process {NARX} models},
  author={Worden, K. and Surace, C. and Becker, W.},
  journal={Procedia {E}ngineering},
  volume={199},
  pages={1994--2000},
  year={2017},
  publisher={Elsevier}
}

@book{Bishop,
	author      = {C.M.\ Bishop},
	title       = {Neural Networks for Pattern Recognition},
	year        = {1995},
	publisher   = {Oxford University Press}
}

@article{worden2025machine,
  title={Machine-learning perspectives on {V}olterra system identification},
  author={Worden, K. and Rogers, T. and Preston, O.},
  journal={Philosophical {T}ransactions of the {R}oyal {S}ociety {A}: {M}athematical, {P}hysical and {E}ngineering {S}ciences},
  volume={383},
  number={2305},
  year={2025},
  publisher={The {R}oyal {S}ociety}
}

@article{kipf2016variational,
  title={Variational graph auto-encoders},
  author={Kipf, T.N. and Welling, M.},
  journal={arXiv preprint arXiv:1611.07308},
  year={2016}
}

@article{Kipf,
    title     = {Semi-supervised classification with graph convolutional networks},
    author    = {T.N.\ Kipf and M.\ Welling},
    journal   = {arXiv preprint:1609.02907},
    year      = {2016}
}

@article{wu2018moleculenet,
  title={Molecule{N}et: a benchmark for molecular machine learning},
  author={Wu, Z. and Ramsundar, B. and Feinberg, E.N. and Gomes, J. and Geniesse, C. and Pappu, A.S. and Leswing, K. and Pande, V.},
  journal={Chemical science},
  volume={9},
  number={2},
  pages={513--530},
  year={2018},
  publisher={Royal Society of Chemistry}
}

@article{tsialiamanis2021foundations,
  title={Foundations of population-based {SHM}, {P}art {IV}: {T}he geometry of spaces of structures and their feature spaces},
  author={Tsialiamanis, G. and Mylonas, C. and Chatzi, E. and Dervilis, N. and Wagg, D.J. and Worden, K.},
  journal={Mechanical {S}ystems and {S}ignal {P}rocessing},
  volume={157},
  pages={107692},
  year={2021},
  publisher={Elsevier}
}

@article{PBSHM5MSSP,
  title={Foundations of population-based {SHM}, {P}art {V}: {N}etwork, framework and database},
  author={Brennan, D.S. and Gosliga, J. and Cross, E.J. and Worden, K.},
  journal={Mechanical {S}ystems and {S}ignal {P}rocessing},
  volume={223},
  pages={111602},
  year={2025},
  publisher={Elsevier}
}

@article{duthe2023graph,
  title={Graph neural networks for aerodynamic flow reconstruction from sparse sensing},
  author={Duth{\'e}, G. and Abdallah, I. and Barber, S. and Chatzi, E.},
  journal={arXiv preprint arXiv:2301.03228},
  year={2023}
}

@article{haywood2026piggo,
  title={{PiGGO}: Physics-Guided Learnable Graph Kalman Filters for Virtual Sensing of Nonlinear Dynamic Structures under Uncertainty},
  author={Haywood-Alexander, M. and Duth{\'e}, G. and Chatzi, E.},
  journal={arXiv preprint arXiv:2604.26593},
  year={2026}
}

@article{kingma2014adam,
  title={Adam: {A} method for stochastic optimization},
  author={Kingma, D.P. and Ba, J.},
  journal={arXiv preprint arXiv:1412.6980},
  year={2014}
}

@article{avendano2021virtual,
  title={Virtual fatigue diagnostics of wake-affected wind turbine via {G}aussian {P}rocess {R}egression},
  author={Avenda{\~n}o-Valencia, L.D. and Abdallah, I. and Chatzi, E.},
  journal={Renewable Energy},
  volume={170},
  pages={539--561},
  year={2021},
  publisher={Elsevier}
}

% If BibTeX is not used, you can enter each bibitem separately using the procedure below, which complies with the IEEE reference style. (When you have less than 10 references, use: "\begin{thebibliography}{1}". Switch to "\begin{thebibliography}{10}" is you have 10 or more references.)

	%\begin{thebibliography}{1}
	%	
	%	\bibitem{heylen1997modal}
	%	W. Heylen, S. Lammens and P. Sas,
	%	\textit{Modal Analysis Theory and Testing}.
	%	Leuven, Belgium: Katholieke Universiteit Leuven, Departement Werktuigkunde, 1997.
	%	
	%	\bibitem{sas1995active}
	%	P. Sas, C. Bao, F. Augusztinovicz and W. Desmet,
	%	``Active control of sound transmission through a double panel partition,''
	%	\textit{Journal of Sound and Vibration}, vol. 180, no. 4, pp. 609-625, 1995.
	%	
	%	\bibitem{boonen2001modified}
	%	R. Boonen and P. Sas,
	%	``Modified smith compensation for feedback active noise control in ducts,''
	%	in \textit{Proceedings of the 2001 International Congress and Exhibition on Noise Control Engineering},
	%   The Hague, The Netherlands, 2001, pp. 619-624.
	%\end{thebibliography}

\end{document}